\documentclass[conference]{IEEEtran}
\IEEEoverridecommandlockouts
\usepackage{cite}
\usepackage{amsmath,amssymb,amsfonts}
\usepackage{algorithmic, algorithm}
\usepackage{graphicx}
\usepackage{textcomp}
\usepackage{xcolor}
\def\BibTeX{{\rm B\kern-.05em{\sc i\kern-.025em b}\kern-.08em
    T\kern-.1667em\lower.7ex\hbox{E}\kern-.125emX}}

\begin{document}

\title{Development of Machine Vision Approach for Bolt Identification based on its Dimension and Pitch for Mechanical Assembly Line
}

\author{
    \IEEEauthorblockN{
    Toshit Jain$^{1,a,*}$,
    Faisel Mushtaq$^{1,a}$,
    K Ramesh$^{2,a}$,
    Sandip Deshmukh$^{2,a}$,
    Tathagata Ray$^{1,a}$,
    Chandu Parimi$^{3,a}$\\
    Praveen Tandon$^{4}$,
    Pramod Kumar Jha$^{4}$}
    \IEEEauthorblockA{$^{1}$Department of Computer Science \& Information Systems, BITS Pilani, Hyderabad Campus
    }
    \IEEEauthorblockA{$^{2}$Department of Mechanical Engineering, BITS Pilani, Hyderabad Campus
    }
    \IEEEauthorblockA{$^{3}$Department of Civil Engineering, BITS Pilani, Hyderabad Campus
    }
    \IEEEauthorblockA{$^{4}$Centre for Advanced Systems, The Defence Research and Development Organisation
    }
    \IEEEauthorblockA{$^{a}$Birla Institute of Technology \& Science, Pilani, Hyderabad Campus
    }
    
    \thanks{$^{*}$Corresponding author

    \textit{Email addresses:}

    Toshit Jain: \texttt{f20170201h@alumni.bits-pilani.ac.in},

    Faisel Mushtaq: \texttt{csefaisel@gmail.com},

    Tathagata Ray: \texttt{rayt@hyderabad.bits-pilani.ac.in}

    }

}

\maketitle

\begin{abstract}
In this work, a highly customizable and scalable vision-based system for the automation of classifying components(bolts in specific) in mechanical assembly lines is described.
The proposed system calculates the required features to classify and identify the different kinds of bolts in the assembly line. The system describes a novel method of calculating the pitch of the bolt in addition to bolt identification and calculating the dimensions of the bolts.
Unlike machine learning-based systems, this identification and classification system is extremely lightweight and can be run on bare minimum hardware such as Raspberry Pi. The system is very fast in the order of milliseconds; hence the system can be used successfully even if the components are steadily moving on a conveyor.
The results show that our system can correctly identify the parts in our dataset with 98\% accuracy using the calculated features.
\end{abstract}


\section{Introduction}
Whenever the production of mechanical items is concerned, assembly lines are the most common production methods. Assembly lines include many small parts like bolts, screws, fasteners, etc. Sorting them and classifying these parts is a challenge. Historically this has been done manually, which is manpower intensive and suffers manual errors. A small inaccuracy can result in significant problems in quality and safety, especially in aerospace vehicle assembly lines. These require strict quality assurance. Hence, introducing a streamlined and automated process for this task holds merit and scope.

We are given various components for aerospace vehicle assembly. These components include different types of bolts, fasteners, nuts, and washers. Bolts have the highest count among all the components. Our objective is to identify and classify bolts - quickly and accurately - from their image on the assembly line. We also aim to provide different features relevant to the assembly process, such as Diameter, Major Axis Length, Pitch, and the Threading Type of the bolt.
We are given thirty-three different kinds of bolts that we aim to classify. The methodology discussed in this paper applies to any type of bolt.

\begin{figure}
    \centering
    \includegraphics[width = \linewidth]{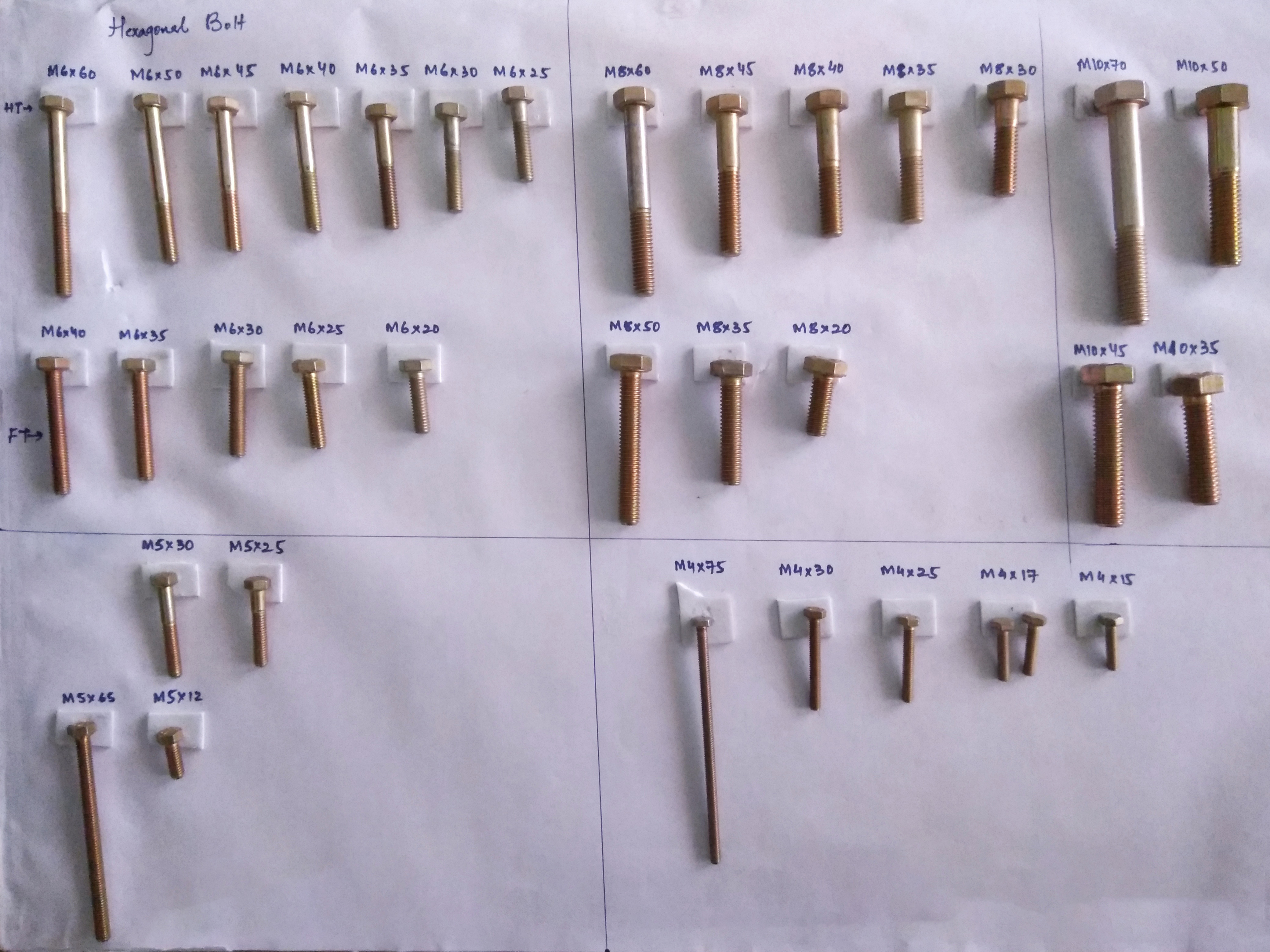}
    \caption{All 33 the Bolts used for this study}
    \label{bolts}
\end{figure}

This paper solely relies on image processing and feature extraction to identify the components, which has certain advantages over the already existing machine learning algorithms \cite{mehrotra1990,canali2014,PERNG2020}. This sole reliance on image processing makes the system lightweight and quick to identify the components without the added expense of training time on higher-end processors such as GPUs. The proposed system can be run on a Raspberry Pi with fast execution times, and the parts can be accurately identified even when moving swiftly on a conveyor. In the case of the introduction of a new kind of bolt in the assembly process, the proposed system is more resilient than the systems which require retraining. This paper also introduces a novel method for calculating the pitch of a bolt precisely using high-resolution cameras.

The paper is organized as follows: a summary of the previous related works are presented in section \ref{LR}. A brief description of the dataset we use is given in section \ref{DS}. The proposed methodology is discussed in section \ref{M} and the results are shown in section \ref{R}. Section \ref{C} concludes the findings of the paper.

\section{Literature Review}\label{LR}
The usage of mechanical components is common in assembly lines, yet the research in automation of recognizing these components is still in its initial stages. Identification of these mechanical components is an important phase for ensuring the correctness of the assembly process. Among some related works, Chávez et al. \cite{chavez2011} proposed a Vision-Based Detection and Labelling of Multiple Vehicle Parts to identify various parts in a car in ordinary scenes. Their study is based on three steps a) Implementation of a machine vision system to identify multiple car parts b) Introduction of a statistical model for elimination of false detections and to infer the possible positions of undetected parts. c) Introduction of geometrical models to model spatial relationships among regions of interest and feasible search zones to delimit search areas. Their system has achieved a precision of 95\% in outdoor scenes. 
A system for geometric feature inspection of mechanical parts \cite{MARSHALL1991} was developed by Marshall et al. using laser range finder data was discussed in 1991.
Among other significant works, Johan et al. \cite{johan2011} proposed a system for recognition of bolt and nut based on backpropagation neural networks. Their system has achieved recognition accuracy of 92\% on a conveyor belt run at a speed of 9 cm/sec. Similarly, Canali et al. \cite{canali2014} proposed a feature-based automatic assembly parts detection and grasping system that can swiftly detect parts and extract useful features by using contour mappings and feature extraction. This work does not identify the parts but detects them with high speed and accuracy without any machine learning methods. In another similar work, Hao et al. (2015) \cite{guo2015} presented a machine vision system for classification of machine parts in real-time by using an embedded image card with Field Programming Gate Array (FPGA) for accelerating the computation. Similarly, Guo et al. \cite{guo2020} also proposed a method for material sorting system based on machine vision that relied heavily on image preprocessing and edge detection but was only effective for a few predefined shapes.
In other similar works, Huang et al. \cite{huang2019} proposed component assembly inspection based on mask R-CNN and support vector machines. Also, Killing et al. (2009) \cite{killing2009} developed a system for the detection of missing fasteners on steel stampings based on neuro-fuzzy and threshold-based picture classification algorithms. The authors conclude that neuro-fuzzy has a far lower RMS error than the threshold-based classification algorithm.

\section{Setup and Dataset}\label{DS}

\begin{figure}
    \centering
    \includegraphics[width = \linewidth]{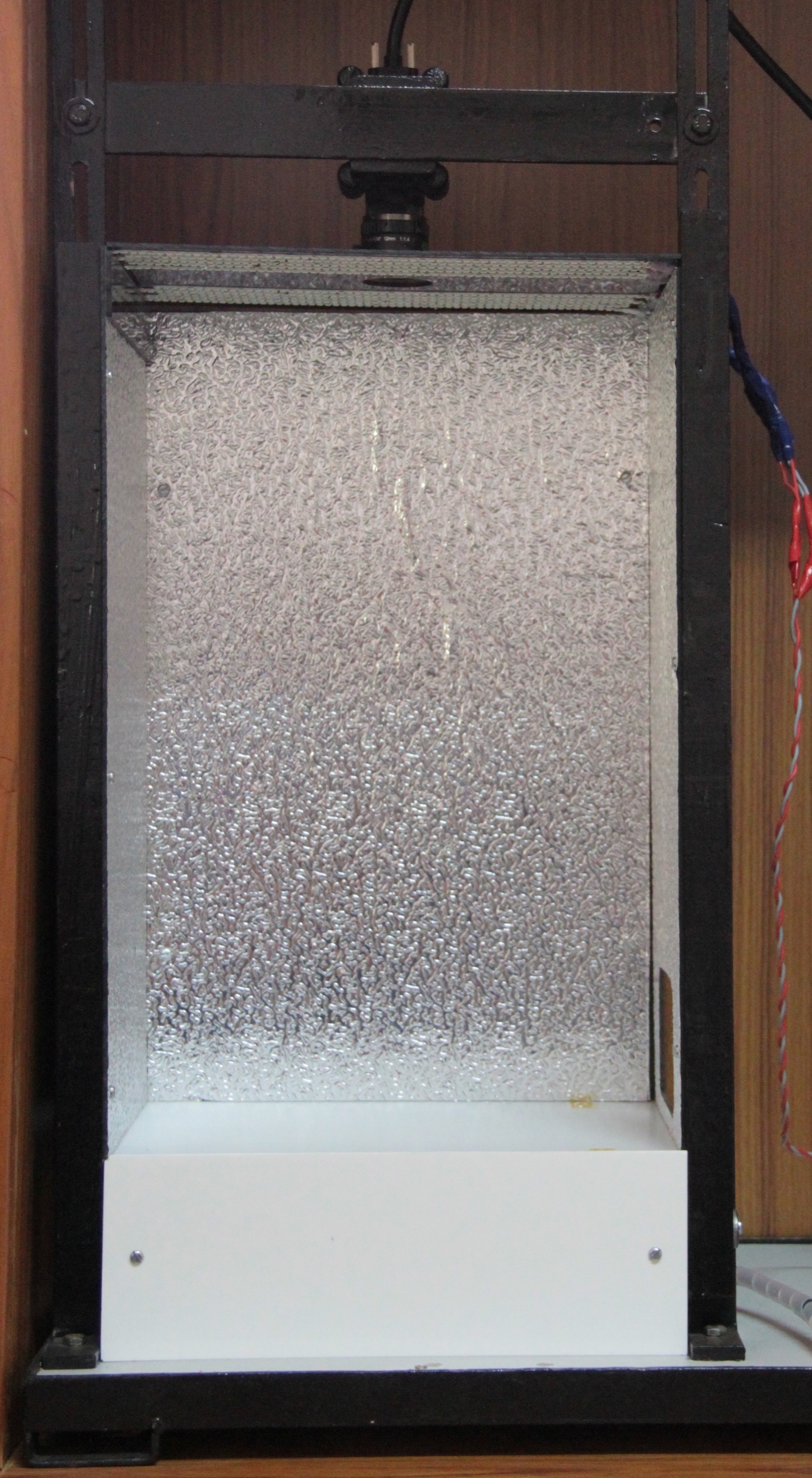}
    \caption{The Setup}
    \label{setup}
\end{figure}

\begin{figure}
    \centering
    \includegraphics[width = \linewidth]{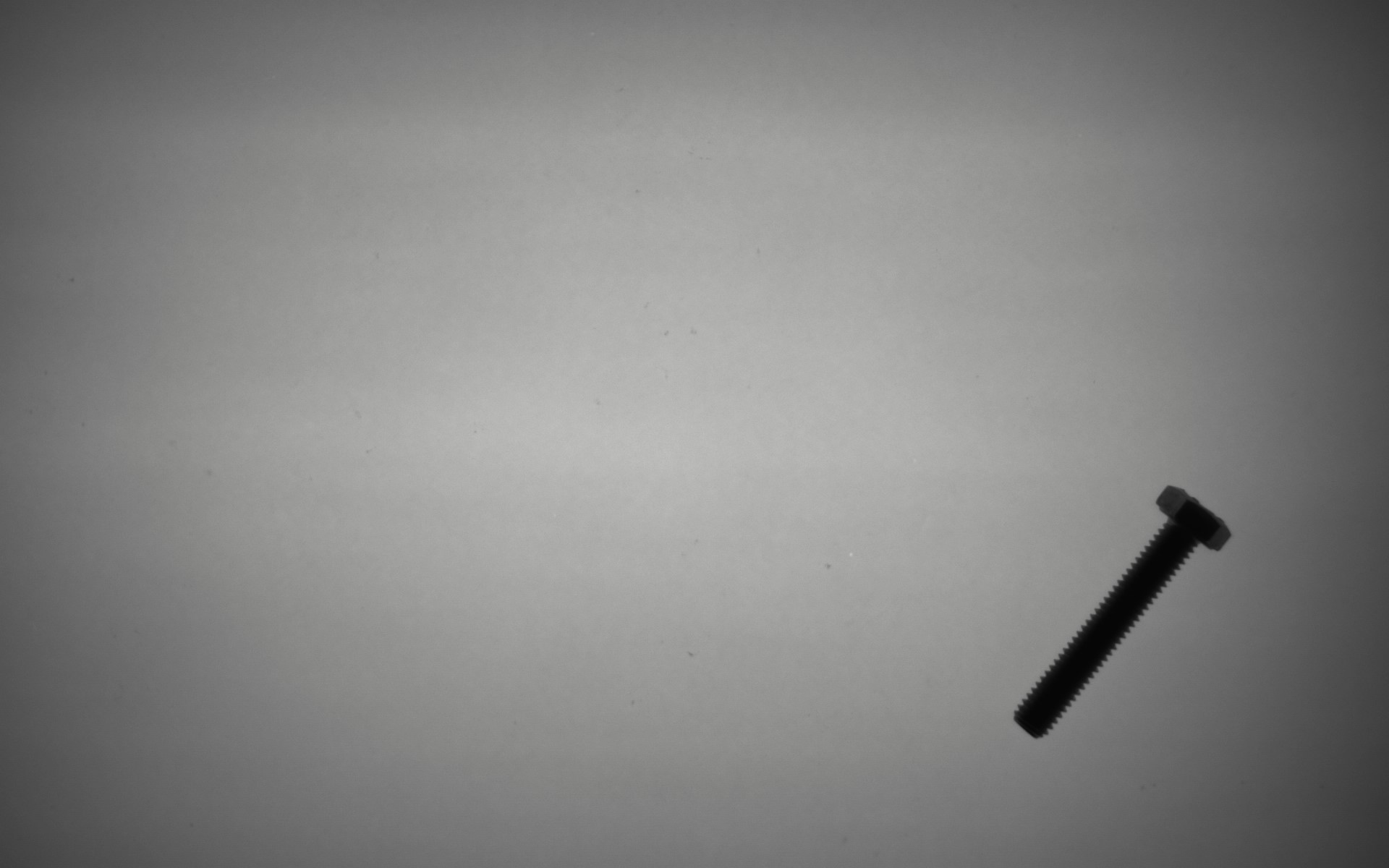}
    \caption{Sample image of a top view in the dataset}
    \label{sample}
\end{figure}

\begin{figure}
    \centering
    \includegraphics[width = \linewidth]{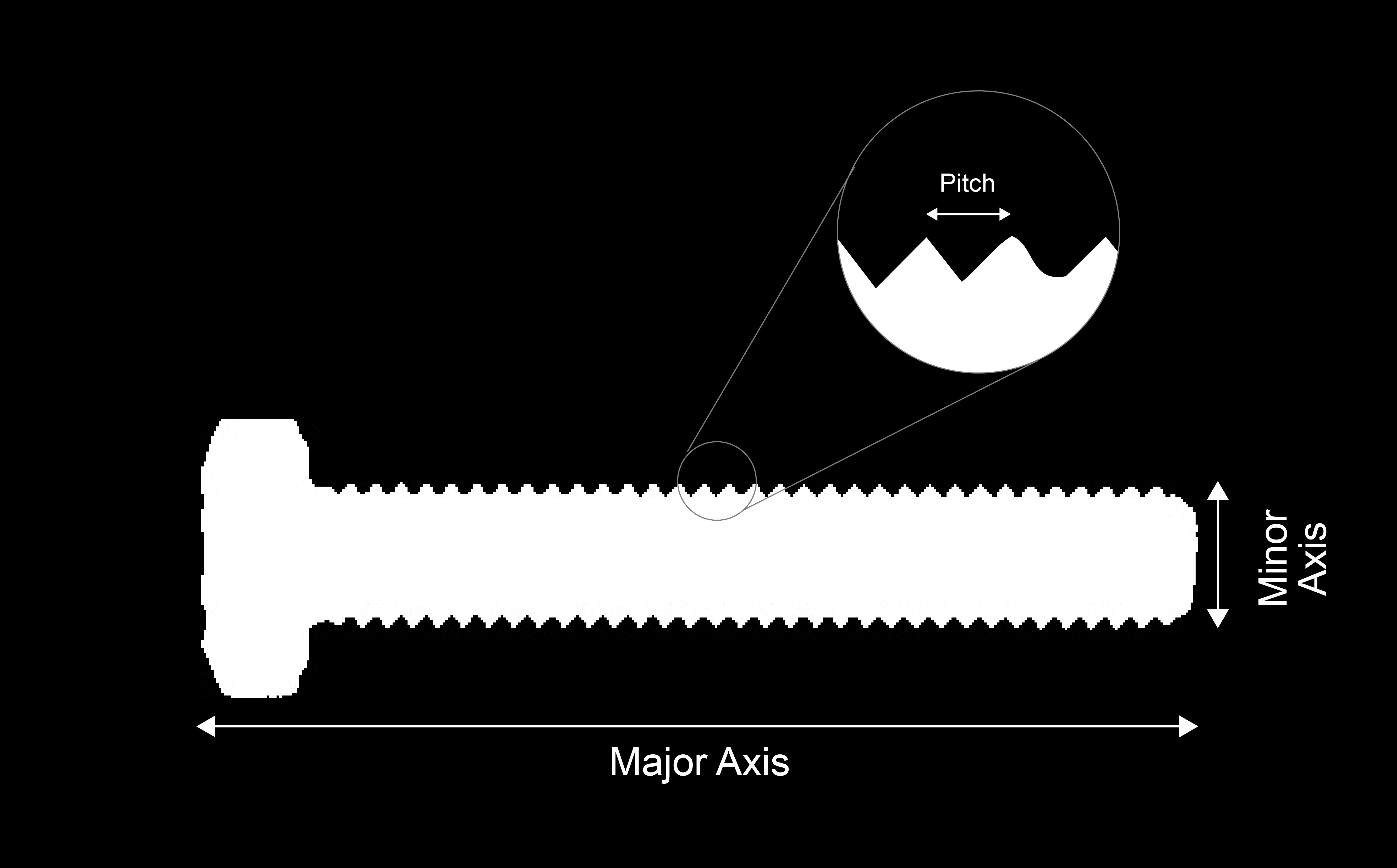}
    \caption{Binary Image of a full thread bolt}
    \label{binary}
\end{figure}

An image acquisition system has been developed to identify bolts for capturing quality images. An opal-frosted acrylic sheet has been used to fabricate the testbed base, supported by a backlit LED source with adjustable light intensity to eliminate shadow and reflection. To capture images, a 4MP monochrome industrial camera has been mounted above the testbed at a fixed position to capture the lateral view of bolts. This lateral view (as shown in Fig. \ref{sample}) of the components is eventually used to determine the dimension and pitch of the components. The setup is also enclosed with black acrylic sheets to block unnecessary light from the surroundings. Since the captured images are the silhouettes of the bolts, they are easy to convert to a binary format using thresholding. We have used the OpenCV library\cite{opencv} \verb|cv2.threshold()| to get the binary equivalent of the image. A binary image is essentially one with just two values: 0 and 1, where value 0 represents black and value 1 represents white. An example of this is shown in Fig. \ref{binary}. Also, since the distance between the camera and the testbed is always constant, we can easily convert the pixels into millimeters by multiplying them with a conversion factor (pixels per mm) to compensate for perspective shortening.

\section{Methodology}\label{M}
\begin{figure}
    \centering
    \includegraphics[width = \linewidth]{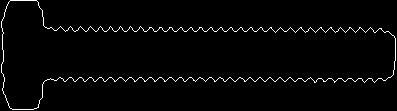}
    \caption{Contour of a bolt}
    \label{contour}
\end{figure}

\begin{figure}
    \centering
    \includegraphics[width = \linewidth]{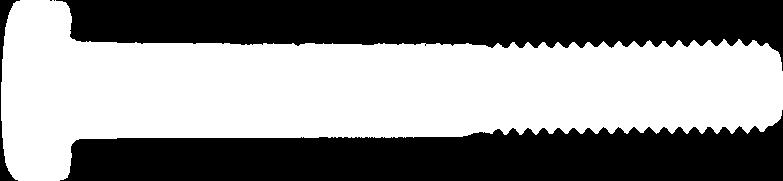}
    \caption{Binary image of a half thread bolt}
    \label{half}
\end{figure}

A bolt is kept inside the setup, and the top camera captures an image. The bolt is identified by three features, namely, the length of the major axis, the length of the minor axis, and the type of threading. Pitch is also an important feature that we calculate.
The other features we focus on are area and perimeter. These two features aren't physically significant, but we use them to calculate other features.
The features are defined as follows:
\begin{itemize}
    \item Length of major axis: This is defined as the full lateral length of the bolt. See Fig. \ref{binary}
    \item Length of minor axis: This is the diameter of the body of the bolt. See Fig. \ref{binary}
    \item Type of threading: We have two types of bolts in our dataset: Full thread bolts and half thread bolts. Fig. \ref{minarea} shows the silhouette of a full thread bolt and Fig. \ref{half} shows the silhouette of a half thread bolt.
    \item Pitch: Pitch is the distance between two consecutive threads. See Fig. \ref{binary}
    \item Area: This is the area of the lateral cross-section of a bolt. The total number of white pixels in Fig. \ref{minarea} is the area of the bolt.
    \item Perimeter: This is the perimeter of the shape that the silhouette of the bolt makes. The total number of white pixels in Fig. \ref{contour} is the perimeter of the bolt.
\end{itemize}

\subsection{Getting the region of interest in minimum possible area}\label{roi}
In a binary image, we only have two types of pixels, black and white, with values 0 and 255 respectively. Getting an upright bounding box is fairly straightforward since the pixels for a bolt form connected components in a graph. The top, right, bottom, and left bounds are determined by the topmost, rightmost, bottom-most, and leftmost white pixels. The \verb|cv2.boundingRect()| method can be used to get this bounding box. Similarly, edge detection can also be done by using \verb|cv2.findContours()|. We find a rectangle, possibly rotated, with the minimum area that encloses the whole component. This is easily implemented by first finding the contours of the bolt by using \verb|cv2.findContours()| method from the OpenCV Library which basically stores the shape of the component and then we get the minimum area rectangle by calling \verb|cv2.minAreaRect()|.

\subsection{Calculation of area and perimeter}\label{areaperimeter}
The area of a component in a binary image is the number of white pixels since the white pixels represent the cross-section of the bolt. The area calculation is very helpful in orienting the image, getting the type of threading, and calculating the pitch. Perimeter is calculated very easily using \verb|cv2.arcLength()| on the contour of the screw since we already have calculated the contours in section \ref{roi}. The contour of a screw looks as shown in Fig. \ref{contour}. The perimeter helps identify the threading type.

\subsection{Aligning and orienting the minimum area rectangle}\label{warping}

We align the minimum area rectangle along the x and y axes obtained in section \ref{roi} by using \verb|cv2.getPerspectiveTransform()| and warping this perspective to an upright rectangle that is aligned with the axes, with the same height and width that was returned by \verb|cv2.minAreaRect()| in section \ref{roi} by using the \verb|cv2.warpPerspective()| method. Then we rotate the aligned rectangle such that the length of the rectangle is longer than its height. This length is the \textbf{length of the major axis}.

\begin{figure}
    \centering
    \includegraphics[width = \linewidth]{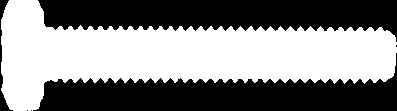}
    \caption{The aligned and oriented minimum area rectangle}
    \label{minarea}
\end{figure}

As a convention, we keep the head of the bolt on the left side of the rectangle. Since our rectangle is already laid out horizontally, we need to determine if the head is on the left or right side. For this, we cut the image from the middle and divided it into two parts. We calculate the areas in both the image's left and right half. Since the image is binary, the area is the total number of white pixels. The head is on the side where we have a greater value of the area. If the area of the right half of the image is larger, then we rotate the image 180 degrees. The image after orienting is shown in Fig. \ref{minarea}.

\subsection{Returning the major and minor axes}\label{axes}
The length of the minimum area rectangle is the length of the major axis, in pixels. To get the length of the minor axis, we cut the oriented image in half and drew a minimum area rectangle on the right half. The width of this new rectangle is \textbf{the length of minor axis}, \textbf{\textit{d}}, in pixels.

\subsection{Removing the head}\label{rh}
We have an oriented binary image of the bolt with its head on the left. In all the types of bolts in our dataset, along the length of the bolt, the head does not take more than 20\% of the length of the bolt. The algorithm to remove the head is similar to binary search. Suppose the length of the image, i.e. the length of the bolt is \textit{l}, and the width of the image, i.e. the width of the head of the bolt is \textit{w}. We also already have the length of the minor axis, \textit{d} An element for our “binary search” algorithm is defined as a one-pixel wide and \textit{w} pixels long strip of our image. The starting element is the leftmost strip of the image and the last element is the strip at \textit{0.2l} from the left end of the image.

In every step of the search, we slice the image from the middle point, \textit{m} to the end of the image, \textit{l}, and get a minimum area rectangle of this sliced image. If the width of this minimum area rectangle, \textit{w'} is approximately the same as \textit{w}, then we repeat this in the right half. If \textit{w'} is approximately the same as \textit{d} then that means the entire head is to the left of \textit{m} and we search again in the left half. We stop when we find an element \textit{h} where the width of the minimum area rectangle from \textit{0} to \textit{h-1} is approximately \textit{w} and the width of the minimum area rectangle from \textit{h} to \textit{l} is approximately \textit{d}. We finally return an image slice from \textit{h} to \textit{l} which corresponds to the body of the bolt.

\textit{Note that} if the head of the bolt is required then we can return the image slice from \textit{0} to \textit{h-1}.

\begin{figure}
    \centering
    \includegraphics[width = \linewidth]{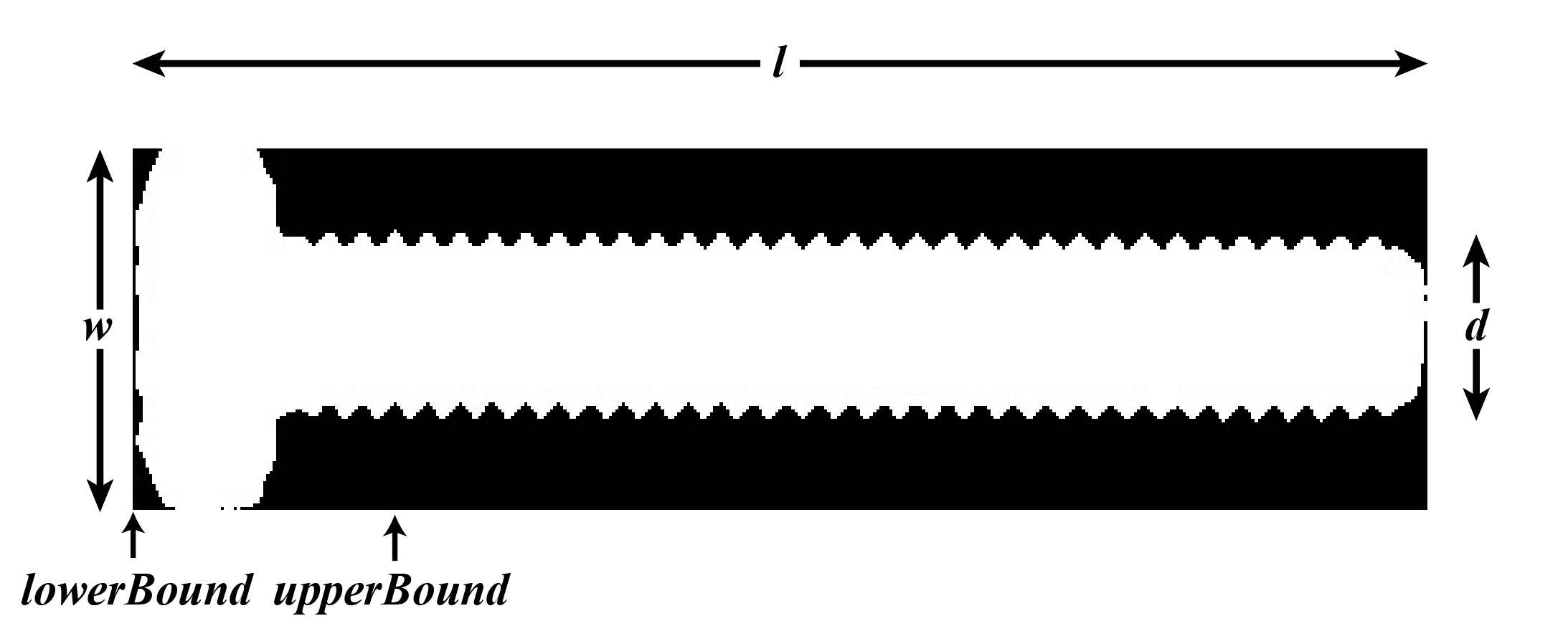}
    \caption{Variables used in algorithm \ref{headlessalgo}}
    \label{lab}
\end{figure}

\begin{figure}
    \centering
    \includegraphics[width = \linewidth]{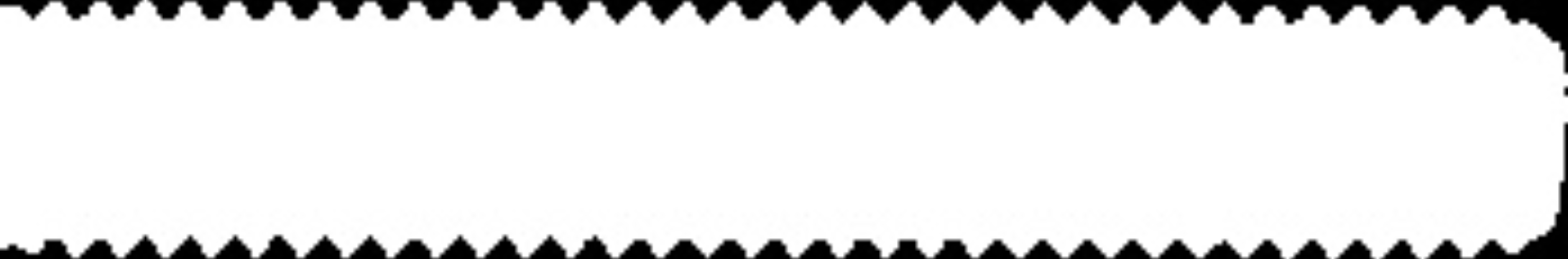}
    \caption{Bolt after it's head is removed (Output of algorithm \ref{headlessalgo})}
    \label{headless}
\end{figure}

Algorithm \ref{headlessalgo} gives this pseudocode in detail. The variables used are visually depicted in Fig. \ref{lab} It uses a parameter $thresh$. This parameter can be set to 0 to get the exact cutting point. In practice, we set this parameter to 5. Though this does not provide the exact cutting point and actually takes away 5 extra strips from the head side of the bolt, this is practically better for the algorithm. This parameter's significance is explained more in section \ref{R}. The output of this algorithm is shown in Fig. \ref{headless}

\begin{algorithm}
    \caption{Algorithm for removing head from a bolt}
    \begin{algorithmic}[1]
    \renewcommand{\algorithmicrequire}{\textbf{Input:}}
    \renewcommand{\algorithmicensure}{\textbf{Output:}}
    \REQUIRE Oriented minimum area rectangle image \textit{img} of the bolt
    \ENSURE  Image of the bolt without the head of the bolt
     \STATE $l \gets length(img)$
     \STATE $w \gets width(img)$
     \STATE $d \gets diameterOfBolt$
     \STATE $lowerBound \gets 0$
     \STATE $upperBound \gets 0.2l$
     \STATE $thresh \gets 5$
     \WHILE{True}
     \STATE $mid \gets (lowerBound + upperBound)/2$
     \STATE $temp \gets minAreaRect(img[mid:l])$
     \STATE $w' \gets width(temp) $
     \IF {($abs(w'-w) \le abs(w'-d)$)}
     \STATE $upperBound \gets mid+1$
     \ELSE 
     \STATE $lowerBound \gets mid-1$
     \STATE $\_temp \gets minAreaRect(img[mid-1:l])$
     \STATE $w'' \gets width(\_temp) $
     \IF {($abs(w''-w) \le abs(w''-d)$)}
     \RETURN $minAreaRect(img[mid+thresh:l])$
     \ENDIF
     \ENDIF
     \ENDWHILE
     \RETURN $minAreaRect(img[mid+thresh:l])$
    \end{algorithmic}
    \label{headlessalgo} 
\end{algorithm}

\subsection{Getting the threading type}
The labeling method that we use to identify a bolt uses three labels: length of the bolt, diameter, and type of threading. Generally, there are two types of threading in the bolts:
\begin{itemize}
    \item Full thread, in which the threading goes all the way from the bottom to the head along the body of the bolt.
    \item Half thread, in which the threading starts from the bottom to about 35-40\% of the length of the bolt.
\end{itemize}
To identify the type of threading, we start with the headless image of the bolt obtained in section \ref{rh}. We divide the image into two parts by cutting it from the middle. Now we compare the two halves to determine if the bolt is half threaded. In a half-threaded bolt, the entire threaded part lies in the right half. We calculate the contours of both halves using the \verb|cv2.findContours()| method in the OpenCV library. If the bolt is half threaded, then the contour of the left half of the bolt is basically a rectangle, which is a convex polygon. So if the \verb|cv2.isContourConvex()| function returns \textit{True} on the left contour, then we conclude the bolt is half threaded. If it returns \textit{False} then we must run some other tests before concluding the bolt to be fully threaded because some noise in the image could have resulted in minor irregularities in the contour that may have resulted it to test negative on the convex polygon check. Then we compare the areas and the perimeters of the two halves. If the perimeter of the right half is considerably larger than that of the left half, then we can conclude that the bolt is half threaded because a fully threaded bolt would have similar perimeters in both halves. Otherwise, we check if the area of the left half is comparable to the area of the whole left part of the image. If this turns out to be true then we can say that the bolt is half threaded. If all the three tests turn out negative only then we say that the bolt is a full thread bolt.

\subsection{Identification of the bolt}\label{id}
The actual size of a bolt differs from the size that is indicated by the name of the bolt. For example, the bolt indicated by M5X12\_FT has a diameter equal to 4.90 mm and a length equal to 12.06 mm. Moreover, a slight variation in the dimensions is observed when the bolt is kept in the middle of the setup, i.e. directly below the camera, as compared to when it is kept in the corner. This is due to perspective shortening. If we directly convert the lengths of the axes that we obtained in section \ref{axes}, from pixels to millimeters, then we must keep these two shortcomings in mind. To overcome these two obstacles, and simplify the problems,  we adopt a different approach for bolt identification. We create a minimum area rectangle of all bolts and store them in a lookup table. These minimum area rectangles are manually verified to be correct and precise. Then, their widths and heights, in pixels, are stored in tabular form in a lookup table. Then, our method calculates the dimensions and the threading type, and the closest match from the lookup table is returned. We use Euclidean distance to calculate the closest match.

\subsection{Calculation of pitch}
Pitch is the distance between two consecutive threads. To calculate pitch, we take the headless image of the bolt obtained in section \ref{rh}. Regardless of the bolt's threading type, 30\% of the screw from the right will always be threaded. We slice out this part of the bolt. We’ll calculate the pitch on this part of the image. First, we construct a minimum area rectangle on this image and align and orient it horizontally as in section \ref{warping}. Next, we calculate its contour and generate its perimeter shape. We also generate the shape of the minimum area rectangle. Ideally, all the crests should touch the upper side of the rectangle. We could calculate the points of intersection of the two contours, and we should be able to calculate pitch. But there may be minor irregularities due to noise, image resolution, blur, etc., and some crests may end up a pixel or two below the upper side of the rectangle. Therefore to be sure that all the crests intersect the rectangle, we nudge the upper side down by some pixels. Therefore for every crest, we get two intersection points. Next, we do an AND operation of the two contours. We traverse the length of the upper side and count the number of white pixels we encounter. Let the total number of white pixels encountered be \textit{n}. We also store the location of the first white pixel, \textit{a} that we encounter, and the last white pixel, \textit{b} we encounter. Subtracting these two gives the stretch length in which we counted the pixels. The total number of white pixels is half of our actual count. Therefore, the average pitch value is given by:
\begin{equation}
pitch = \frac{b-a}{\frac{n}{2}}
\end{equation}

\section{Results}\label{R}

\subsection{Perspective Shortening}

One of the innate challenges of the setup is perspective shortening. The lengths of the bolt as calculated in the corner can differ from the lengths as calculated in the center by 1.4\% in the worst case. The conversion factor is calculated for the center of the setup and gives more accurate results when the bolt is kept in the center of the setup. For identification, the results are the same irrespective of where the bolt is kept. This is because an error of 1.4\% does not change the class of the bolt. No two components have dimensions within 1.4\% of each other. But this error does become an issue for pitch calculation. Therefore, all the pitch calculations are made with the bolt kept in the center for more accurate results. The differences in measurements for some bolts are shown in table \ref{tab1}. From this table, it is apparent that the bolt measurements (in pixels) appear to be larger when the bolt is kept in the center as compared to when it is kept in the corner in different orientations. The difference is observed to be maximum when the bolt is kept in the extreme corner parallel to the side of the testbed.

\begin{table}[htbp]
    \caption{Measurements of Bolts when kept at the corners and the center, in pixels}
    \begin{center}
    \begin{tabular}{|c|c|c|c|c|c|}
    \hline
    \textbf{Bolt} &\multicolumn{2}{|c|}{\textbf{Center}} &\multicolumn{2}{|c|}{\textbf{Corner}} &\textbf{Percentage} \\
    \cline{2-5}
    \textbf{Name} & \textbf{\textit{Height}} & \textbf{\textit{Width}} & \textbf{\textit{Height}} & \textbf{\textit{Width}} & \textbf{Error} \\
    \hline
    \textbf{M8x35\_HT} & 147 & 407 & 147 & 411 & 0.98\% \\
    \hline
    \textbf{M10x50\_HT} & 181  & 577 & 183 & 579 & 1.22\% \\
    \hline
    \textbf{M10x35\_FT} & 179 &426  & 181 & 428 & 1.21\% \\
    \hline
    \textbf{M4x75\_FT} & 75  & 781 & 75 & 783 & 0.26\% \\
    \hline
    \end{tabular}
    \label{tab1}
    \end{center}
\end{table}

\subsection{Identification of the Bolts}

\begin{figure}
    \centering
    \includegraphics[width = \linewidth]{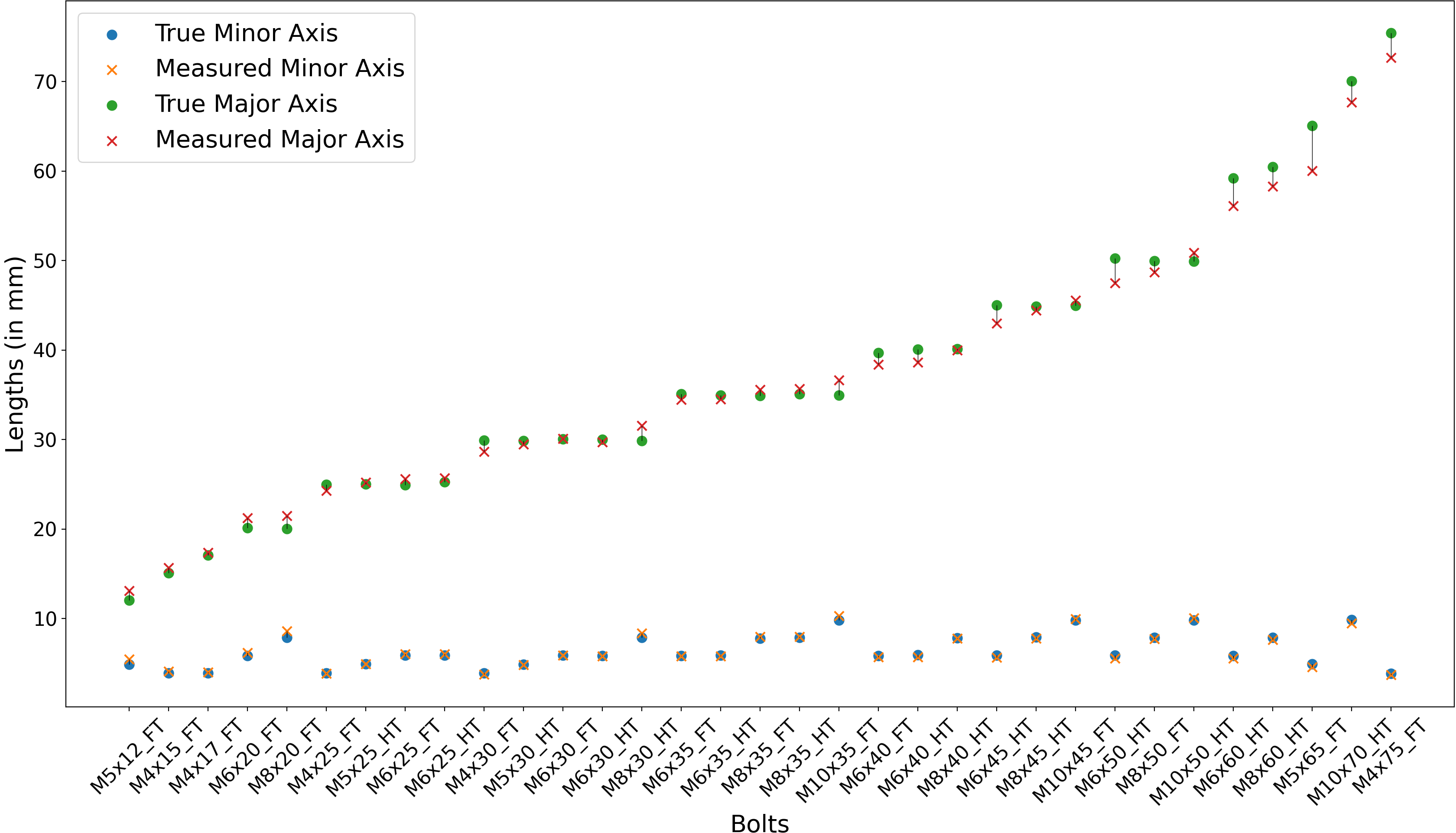}
    \caption{Differences between the experimental and actual measurements}
    \label{plot1}
\end{figure}

\begin{figure}
    \centering
    \includegraphics[width = \linewidth]{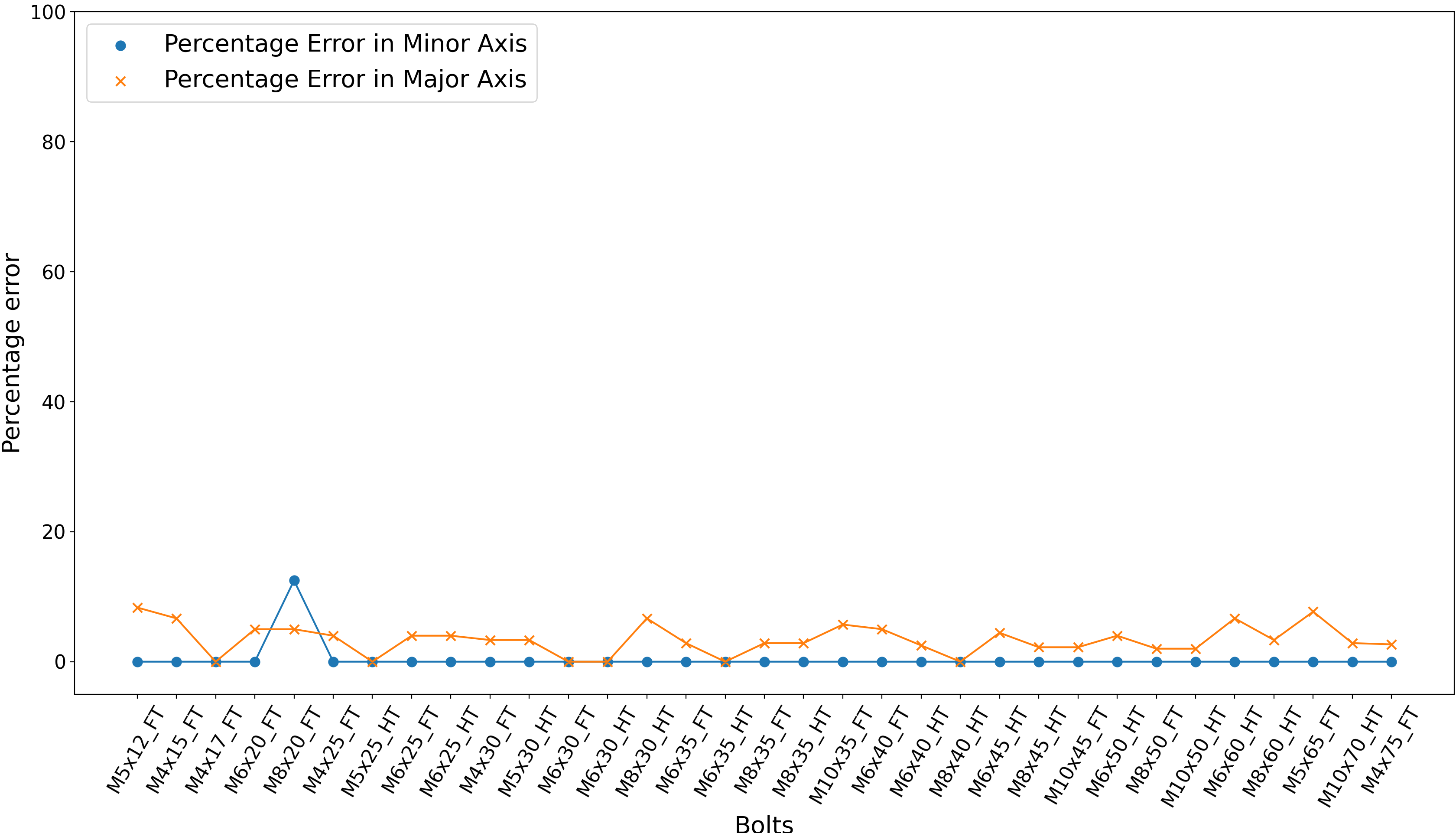}
    \caption{Percentage errors of bolt measurements after rounding off to nearest whole number}
    \label{plot2}
\end{figure}

The major and minor axes of all the bolts were measured. The conversion factor used was 12.42 pixels per millimeter. Both in Fig. \ref{plot1} and Fig. \ref{plot2}, the bolts are arranged on the x-axis in increasing order of the lengths of the bolts.
The difference in the measurements (in millimeters) is shown in Fig. \ref{plot1}. The measurements of the minor axis are very close to the actual values.
We can see a larger difference in measurements of the major axis as the length of the bolts increases. But as seen in Fig. \ref{plot2} the actual percentage error is very less as the length of the bolt is very large.
For example, the length of the last bolt was measured to be 72.70mm while the actual length was 75.43 mm. This error of 2.73mm on 75.43mm bolts is a very low 2.67\% of error.
In Fig. \ref{plot2} the percentage errors in measurements are calculated after rounding off to the nearest integer.
There is a peak in percentage errors in minor axis measurement. The M8x20\_FT bolt was measured to have 8.61 mm minor axis.
This resulted in it to getting rounded off to 9mm and this 1 mm error resulted in a 12.5\% error. But eventually, after taking the major axis into account, the bolt will get classified into the correct class. This is because we use Euclidean distance for identification.

\begin{figure}
    \centering
    \includegraphics[width = \linewidth]{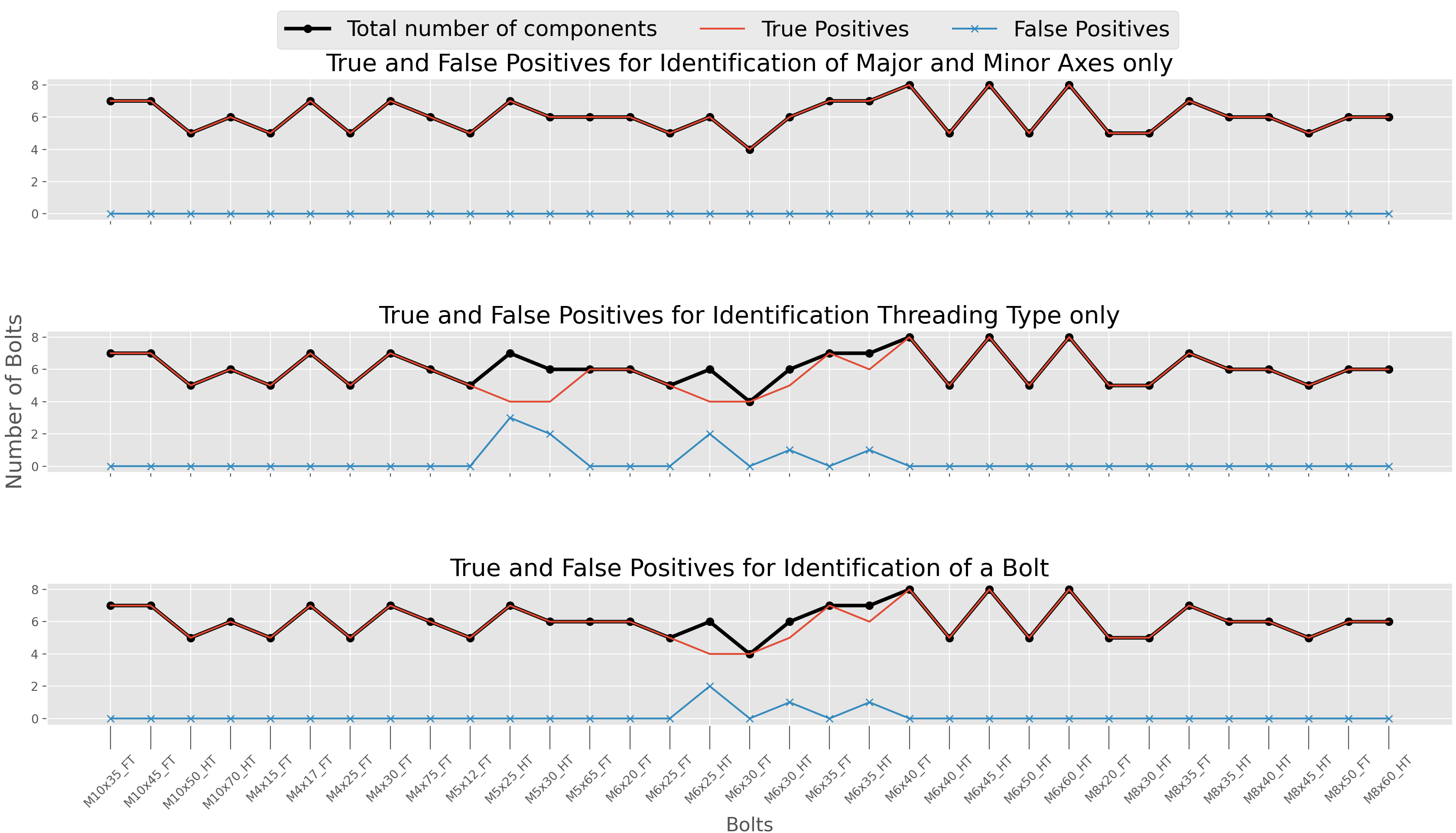}
    \caption{True positives and false positives for identification of different parameters of a bolt over a set of 200 images}
    \label{plot3}
\end{figure}

Fig. \ref{plot3} shows the results of an experiment on 200 bolts. 5 to 8 bolts were placed on the testbed with random orientation. The results were produced by comparing the Euclidean distance of the height and width of the minimum area rectangles with the template table as discussed in section \ref{id}. The first graph in Fig. \ref{plot3} shows that the dimensions of all the 200 bolts were correctly identified with 100\% accuracy. The true positive curve exactly follows the total number of components curve. The false positives are zero. The second graph in Fig. \ref{plot3} shows the results of the identification of the type of threading. A total of 9 false positives were detected. All 9 false positives were half-threaded bolts that were incorrectly identified as full-threaded bolts. The accuracy for identification of threading type remains 95.5\%. The full-threaded bolts were all correctly identified as full-threaded with 100\% accuracy. In the third graph in Fig. \ref{plot3} the results of the identification of the whole bolts are shown. The false positives drop to just 4. This is because if we get a bolt that matches only one entry in the lookup table then the threading type is not needed for the full identification of the bolt. For example, we can see that the M5x25\_HT bolt was incorrectly identified as full-threaded 3 out of 7 times. But these 3 false positives drop to zero as we have no full threaded bolt with the same dimensions. Hence we prefer the dimensional results and ignore the threading result in these cases. The accuracy we get for the identification of the whole bolt is 98\% on a test set of 200 images.

\subsection{Head Removal}

\begin{figure}
    \centering
    \includegraphics[width = \linewidth]{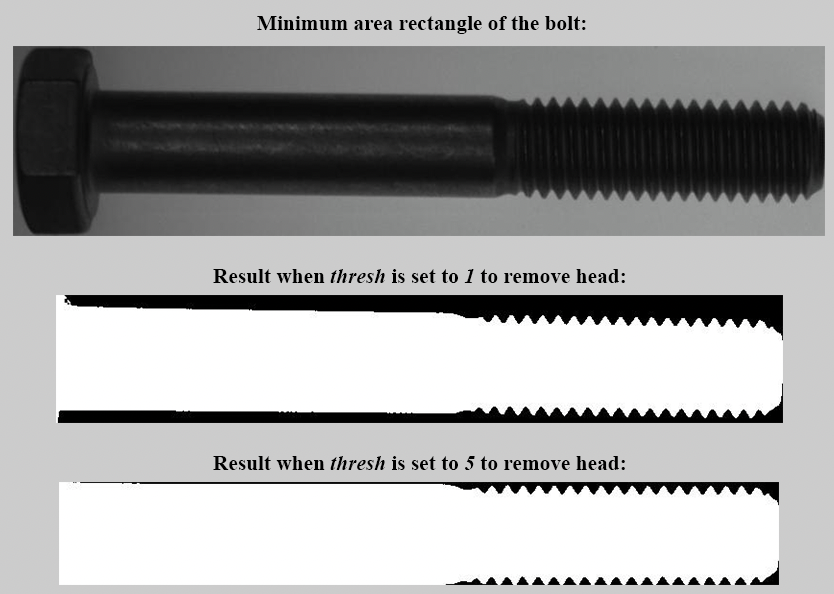}
    \caption{Result of algorithm \ref{headlessalgo} with \textit{thresh} values 1 and 5}
    \label{thresh}
\end{figure}

In section \ref{rh} the parameter \textit{thresh} was used to control the positions of the cut point for the removal of the head from a bolt. When the bolt is kept in the center of the testbed then the value of \textit{thresh} can be kept as 0 and that works perfectly. When the bolt is kept in the corner, the head removed image can look like Fig. \ref{thresh}. Therefore we keep the value of \textit{thresh} as 5 for these corner cases.

\subsection{Pitch Calculation}
The camera used in our setup is 4MP. This resolution limits the amount of details captured. Because of this limitation, the pitch calculation was done only on larger components, and the component was placed in the center of the testbed, directly below the camera. Even though the camera was of low resolution, we were able to get the pitch between $\pm$ 0.07 mm of the actual pitch for all the bolts that we calculated pitch for. This can easily be made more accurate by using cameras with higher resolution or by reducing the distance between the camera and the testbed. 

\subsection{Performance}
The proposed system has shown good performance in identifying different bolts. The system was implemented in Python and run on a) a Macbook with an Intel i5 at 2.30GHz, 8192MiB RAM, and macOS 12, and b) a Raspberry Pi with a quad-core BCM2711 at 1.500GHz, 1872MiB RAM, and Raspbian GNU/Linux 10. Table \ref{tab2} summarizes the execution times of the algorithms in the task of getting all the features, namely, Pitch Calculation, Calculation of the Major and Minor Axes, Threading type Identification, and Classifying the Bolt.

\begin{table}[htbp]
    \caption{Performance of the System}
    \begin{center}
    \begin{tabular}{|c|c|}
    \hline
    \multicolumn{2}{|c|}{\textbf{Average time of execution}} \\
    \hline
    Macbook & 105 ms \\
    \hline
    Raspberry Pi & 233 ms  \\
    \hline
    \end{tabular}
    \label{tab2}
    \end{center}
\end{table}

\section{Conclusion}\label{C}

In this paper, a system for the identification of various features for various bolts was presented. The system has immediate applications in industrial assembly lines that contain a lot of bolts. The system describes a way to get the type of threading, dimensions, and pitch of a bolt. From the experimental results, the method discussed guarantees the successful identification of bolts.
The model was 100\% successful in calculating the lengths of the major and minor axes. This accurate calculation of the dimension was also useful in sorting out defective and/or wrong parts.
The accuracy of the model was found to be 98\% for the correct identification of the bolts.
The identification of bolts that were larger than 40 mm was 100\% accurate. Hence, the accuracy of the model can be improved to 100\% with higher resolution cameras.
The pitch calculation is close to accurate on larger parts but is unreliable for smaller bolts in the current setup due to hardware limitations. Currently, the pitch can be calculated within an error range of $\pm$0.07 mm.
The system is fast and, even with limited computing resources, can successfully identify components on a swiftly moving conveyor belt.

In future works, we intend to define methods for the classification of different components like nuts, bolts, washers, and other fasteners using computer vision, and then propose similar image processing methods for extracting the features of these classified components.

\section*{Acknowledgments}
The authors gratefully acknowledge ARMREB, New Delhi for providing financial support to carry out this research (Sanction Letter No.: ARMREB/ADMB/2020/224). The authors are thankful to Birla Institute of Technology \& Science, Pilani, Hyderabad Campus for their support in carrying out this work.

\bibliographystyle{IEEEtran}
\bibliography{references.bib}

\end{document}